\def\year{2020}\relax
\documentclass[letterpaper]{article} 
\usepackage{aaai20}  
\usepackage{times}  
\usepackage{helvet} 
\usepackage{courier}  
\usepackage[hyphens]{url}  
\usepackage{graphicx} 
\usepackage{multirow}
\usepackage{graphicx}  
\usepackage[lofdepth,lotdepth]{subfig}
\usepackage{amsfonts}
\title{Learning High-order Structural and Attribute information by Knowledge Graph Attention Networks for Enhancing Knowledge Graph Embedding}
\author{Wenqiang Liu, Hongyun cai, Xu Cheng, Sifa Xie ,Yipeng Yu,Hanyu Zhang\\
Interactive Entertainment Group, Tencent Inc, Shenzhen, China.\\
liuwenqiangcs@gmail.com, \{laineycai,alexcheng,sifaxie,dukehyzhang\}\\
@tencent.com, yypzju@163.com\\
}

\begin{document}

\maketitle

\begin{abstract}
The goal of representation learning of knowledge graph is to encode both entities and relations into a low-dimensional embedding spaces. Many recent works have demonstrated the benefits of knowledge graph embedding on knowledge graph completion task, such as relation extraction. However, we observe that: 1) existing method just take direct relations between entities into consideration and fails to express high-order structural relationship between entities; 2) these methods just leverage relation triples of KGs while ignoring a large number of attribute triples that encoding rich semantic information. To overcome these limitations, this paper propose a novel knowledge graph embedding method, named (\textbf{KANE}), which is inspired by the recent developments of graph convolutional networks (GCN). KANE can capture both high-order structural and attribute information of KGs in an efficient, explicit and unified manner under the graph convolutional networks framework. Empirical results on three datasets show that KANE significantly outperforms seven state-of-arts methods. Further analysis verify the efficiency of our method and the benefits brought by the attention mechanism.
\end{abstract}

\section{Introduction}
In the past decade, many large-scale \textit{Knowledge Graphs} (KGs), such as Freebase \cite{bollacker2008freebase}, DBpedia \cite{auer2007dbpedia} and YAGO \cite{suchanek2007yago} have been built to represent human complex knowledge about the real-world in the machine-readable format.  The facts in KGs are usually encoded in the form of triples $(\textit{head entity}, relation, \textit{tail entity})$ (denoted $(h, r, t)$ in this study) through the Resource Description Framework\footnote{https://www.w3.org/TR/rdf11-concepts/}, e.g.,$(\textit{Donald Trump}, Born In, \textit{New York City})$. Figure \ref{Fig:kgs} shows the subgraph of knowledge graph about the family of Donald Trump. In many KGs, we can observe that some relations indicate attributes of entities, such as the $\textit{Born}$ and $\textit{Abstract}$ in Figure \ref{Fig:kgs}, and others indicates the relations between entities (the head entity and tail entity are real world entity). Hence, the relationship in KG can be divided into relations and attributes, and correspondingly two types of triples, namely relation triples and attribute triples \cite{sun2017cross}. A relation triples in KGs represents relationship between entities, e.g.,$(\textit{Donald Trump},Father of, \textit{Ivanka Trump})$, while attribute triples denote a literal attribute value of an entity, e.g.,$(\textit{Donald Trump},Born, \textit{"June 14, 1946"})$. 

\begin{figure}[tp] 
	\centering
	\includegraphics[width=0.5\textwidth]{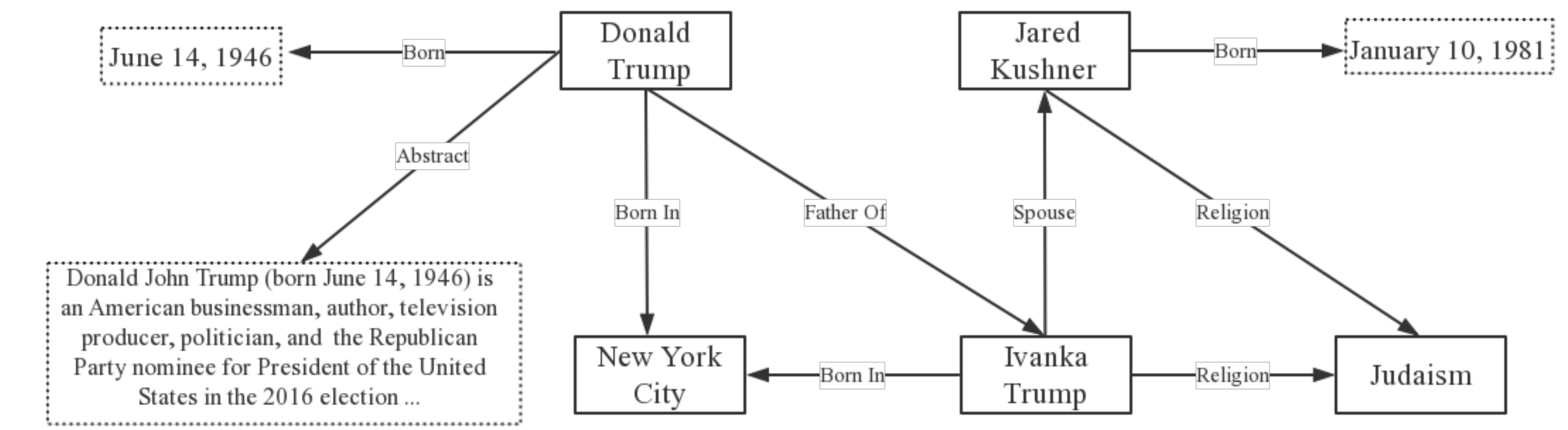}
	\caption{Subgraph of a knowledge graph contains entities, relations and attributes.}
	\label{Fig:kgs}
\end{figure}

Knowledge graphs have became important basis for many artificial intelligence applications, such as recommendation system \cite{wang2018ripplenet}, question answering \cite{hao2017end} and information retrieval \cite{xiong2017explicit}, which is attracting growing interests in both academia and industry communities. A common  approach to apply KGs in these artificial intelligence applications is through embedding, which provide a simple method to encode both entities and relations into a continuous low-dimensional embedding spaces. Hence, learning distributional representation of knowledge graph has attracted many research attentions in recent years. TransE \cite{bordes2013translating} is a seminal work in representation learning low-dimensional vectors for both entities and relations. The basic idea behind TransE is that the embedding $\textbf{t}$ of tail entity should be close to the head entity's embedding $\textbf{r}$ plus the relation vector $\textbf{t}$ if $(h, r, t)$ holds, which indicates $\textbf{h}+\textbf{r}\approx \textbf{t}$. This model provide a flexible way to improve the ability in completing the KGs, such as predicating the missing items in knowledge graph. Since then, several methods like TransH \cite{wang2014knowledge} and TransR \cite{lin2015learning}, which represent the relational translation in other effective forms, have been proposed. Recent attempts focused on either incorporating extra information beyond KG triples \cite{xie2016representation,fatemi2019improved,an2018accurate,guan2019knowledge}, or designing more complicated strategies \cite{ding2018improving,vilnis2018probabilistic,cai2018kbgan}.

While these methods have achieved promising results in KG completion and link predication, existing knowledge graph embedding methods still have room for improvement. First, TransE and its most extensions only take direct relations between entities into consideration. We argue that the high-order structural relationship between entities also contain rich semantic relationships and incorporating these information can improve model performance. For example the fact $\textit{Donald Trump}\stackrel{Father of}{\longrightarrow}\textit{Ivanka Trump}\stackrel{Spouse}{\longrightarrow}\textit{Jared Kushner} $ indicates the relationship between entity \textit{Donald Trump} and entity \textit{Jared Kushner}. Several path-based methods have attempted to take multiple-step relation paths into consideration for learning high-order structural information of KGs \cite{lin2015modeling,toutanova2016compositional}. But note that huge number of paths posed a critical complexity challenge on these methods. In order to enable efficient path modeling, these methods have to make approximations by sampling or applying path selection algorithm. We argue that making approximations has a large impact on the final performance. 

Second, to the best of our knowledge, most existing knowledge graph embedding methods just leverage relation triples of KGs while ignoring a large number of attribute triples. Therefore, these methods easily suffer from sparseness and incompleteness of knowledge graph. Even worse, structure information usually cannot distinguish the different meanings of relations and entities in different triples. We believe that these rich information encoded in attribute triples can help explore rich semantic information and further improve the performance of knowledge graph. For example, we can learn date of birth and abstraction from values of \textit{Born} and \textit{Abstract} about \textit{Donald Trump} in Figure \ref{Fig:kgs}. There are a huge number of attribute triples in real KGs, for example the statistical results in \cite{sun2017cross} shows attribute triples are three times as many as relationship triples in English DBpedia (2016-04)\footnote{http://wiki.dbpedia.org/downloads-2016-04}. Recent a few attempts try to incorporate attribute triples \cite{fatemi2019improved,an2018accurate}. However, these are two limitations existing in these methods. One is that only a part of attribute triples are used in the existing methods, such as only entity description is used in \cite{an2018accurate}. The other is some attempts try to jointly model the attribute triples and relation triples in one unified optimization problem. The loss of two kinds triples has to be carefully balanced during optimization. For example, \cite{sun2017cross} use hyper-parameters to weight the loss of two kinds triples in their models.

Considering limitations of existing knowledge graph embedding methods, we believe it is of critical importance to develop a model that can capture both high-order structural and attribute information of KGs in an efficient, explicit and unified manner. Towards this end, inspired by the recent developments of graph convolutional networks (GCN) \cite{kipf2017semi}, which have the potential of achieving the goal but have not been explored much for knowledge graph embedding, we propose \textbf{K}nowledge Graph \textbf{A}ttention \textbf{N}etworks for \textbf{E}nhancing Knowledge Graph Embedding (\textbf{KANE}). The key ideal of KANE is to aggregate all attribute triples with bias and perform embedding propagation based on relation triples when calculating the representations of given entity. Specifically, two carefully designs are equipped in KANE to correspondingly address the above two challenges: 1) recursive embedding propagation based on relation triples, which updates a entity embedding. Through performing such recursively embedding propagation, the high-order structural information of kGs can be successfully captured in a linear time complexity; and 2) multi-head attention-based aggregation. The weight of each attribute triples can be learned through applying the neural attention mechanism \cite{velivckovic2017graph}.


In experiments, we evaluate our model on two KGs tasks including knowledge graph completion and entity classification. Experimental results on three datasets shows that our method can significantly outperforms state-of-arts methods. 

The main contributions of this study are as follows:

1) We highlight the importance of explicitly modeling the high-order structural and attribution information of KGs to provide better knowledge graph embedding.

2) We proposed a new method KANE, which achieves can capture both high-order structural and attribute information of KGs in an efficient, explicit and unified manner under the graph convolutional networks framework.

3) We conduct experiments on three datasets, demonstrating the effectiveness of KANE and its interpretability in understanding the importance of high-order relations.

\section{Related Work} 
In recent years, there are many efforts in Knowledge Graph Embeddings for KGs aiming to encode entities and relations into a continuous low-dimensional embedding spaces. Knowledge Graph Embedding provides a very simply and effective methods to apply KGs in various artificial intelligence applications. Hence, Knowledge Graph Embeddings has attracted many research attentions in recent years. The general methodology is to define a score function for the triples and finally learn the representations of entities and relations by minimizing the loss function $f_r(h,t)$, which implies some types of transformations on $\textbf{h}$ and $\textbf{t}$. TransE \cite{bordes2013translating} is a seminal work in knowledge graph embedding, which assumes the embedding $\textbf{t}$ of tail entity should be close to the head entity's embedding $\textbf{r}$ plus the relation vector $\textbf{t}$ when  $(h, r, t)$ holds as mentioned in section ``Introduction". Hence, TransE defines the following loss function:
\begin{equation}\label{eq1} 
f_r(h,t)=||\textbf{h}+\textbf{r}-\textbf{t}||_{l_{1}/l_{2}}.
\end{equation}
TransE regarding the relation as a translation between head entity and tail entity is inspired by the word2vec \cite{mikolov2013distributed}, where relationships between words often correspond to translations in latent feature space.  This model achieves a good trade-off between computational efficiency and accuracy in KGs with thousands of relations. but this model has flaws in dealing with one-to-many, many-to-one and many-to-many relations. 

In order to address this issue, TransH \cite{wang2014knowledge} models a relation as a relation-specific hyperplane together with a translation on it, allowing entities to have distinct representation in different relations. TransR \cite{lin2015learning} models entities and relations in separate spaces, i.e., entity space and relation spaces, and performs translation from entity spaces to relation spaces. TransD \cite{ji2015knowledge} captures the diversity of relations and entities simultaneously by defining dynamic mapping matrix. Recent attempts can be divided into two categories: (i) those which tries to incorporate additional information to further improve the performance of knowledge graph embedding, e.g., entity types or concepts \cite{guan2019knowledge}, relations paths \cite{lin2015modeling}, textual descriptions \cite{fatemi2019improved,an2018accurate} and logical rules \cite{guo2016jointly}; (ii) those which tries to design more complicated strategies, e.g., deep neural network models \cite{schlichtkrull2018modeling}.

Except for TransE and its extensions, some efforts measure plausibility by matching latent semantics of entities and relations. The basic idea behind these models is that the plausible triples of a KG is assigned low energies. For examples, Distant Model \cite{bordes2011learning} defines two different projections for head and tail entity in a specific relation, i.e., $\textbf{M}_{r,1}$ and $\textbf{M}_{r,2}$. It represents the vectors of head and tail entity can be transformed by these two projections. The loss function is $f_r(h,t)=||\textbf{M}_{r,1}\textbf{h}-\textbf{M}_{r,2}\textbf{t}||_{1}$. 

Our KANE is conceptually advantageous to existing methods in that: 1) it directly factors high-order relations into the predictive model in linear time which avoids the labor intensive process of materializing paths, thus is more efficient and convenient to use; 2) it directly encodes all attribute triples in learning representation of entities which can capture rich semantic information and further improve the performance of knowledge graph embedding, and 3) KANE can directly factors high-order relations and attribute information into the predictive model in an efficient, explicit and unified manner, thus all related parameters are tailored for optimizing the embedding objective.

\section{Problem Formulation}
In this study, wo consider two kinds of triples existing in KGs: relation triples and attribute triples. Relation triples denote the relation between entities, while attribute triples describe attributes of entities. Both relation and attribute triples denotes important information about entity, we will take both of them into consideration in the task of learning representation of entities. We let $I $ denote the set of IRIs (Internationalized Resource Identifier), $B $ are the set of blank nodes, and $L $ are the set of literals (denoted by quoted strings). The relation triples and attribute triples can be formalized as follows:

\textbf{Definition 1. Relation and Attribute Triples:} A set of Relation triples $ T_{R} $ can be represented by $ T_{R} \subset E \times R \times E $, where $E \subset I \cup B $ is set of entities, $R \subset I$ is set of relations between entities. Similarly, $ T_{A} \subset E \times R \times A $ is the set of attribute triples, where $ A \subset I \cup B \cup L $ is the set of attribute values.

\textbf{Definition 2. Knowledge Graph:} A KG consists of a combination of relation triples in the form of $ (h, r, t)\in T_{R} $, and attribute triples in form of $ (h, r, a)\in T_{A} $.  Formally, we represent a KG as $G=(E,R,A,T_{R},T_{A})$, where $E=\{h,t|(h,r,t)\in T_{R} \cup (h,r,a)\in T_{A}\}$ is set of entities, $R =\{r|(h,r,t)\in T_{R} \cup (h,r,a)\in T_{A}\}$ is set of relations, $A=\{a|(h,r,a)\in T_{A}\}$, respectively.

The purpose of this study is try to use embedding-based model which can capture both high-order structural and attribute information of KGs that assigns a continuous representations for each element of triples in the form $ (\textbf{h}, \textbf{r}, \textbf{t})$ and  $ (\textbf{h}, \textbf{r}, \textbf{a})$, where Boldfaced $\textbf{h}\in \mathbb{R}^{k}$,  $\textbf{r}\in \mathbb{R}^{k}$, $\textbf{t}\in \mathbb{R}^{k}$ and  $\textbf{a}\in \mathbb{R}^{k}$ denote the embedding vector of head entity $h$, relation $r$, tail entity $t$ and attribute $a$ respectively. 

Next, we detail our proposed model which models both high-order structural and attribute information of KGs in an efficient, explicit and unified manner under the graph convolutional networks framework.

\section{Proposed Model}
In this section, we present the proposed model in detail. We first introduce the overall framework of KANE, then discuss the input embedding of entities, relations and values in KGs, the design of embedding propagation layers based on graph attention network and the loss functions for link predication and entity classification task, respectively.
\begin{figure*}[htp] 
	\centering
	\includegraphics[width=1.1\textwidth]{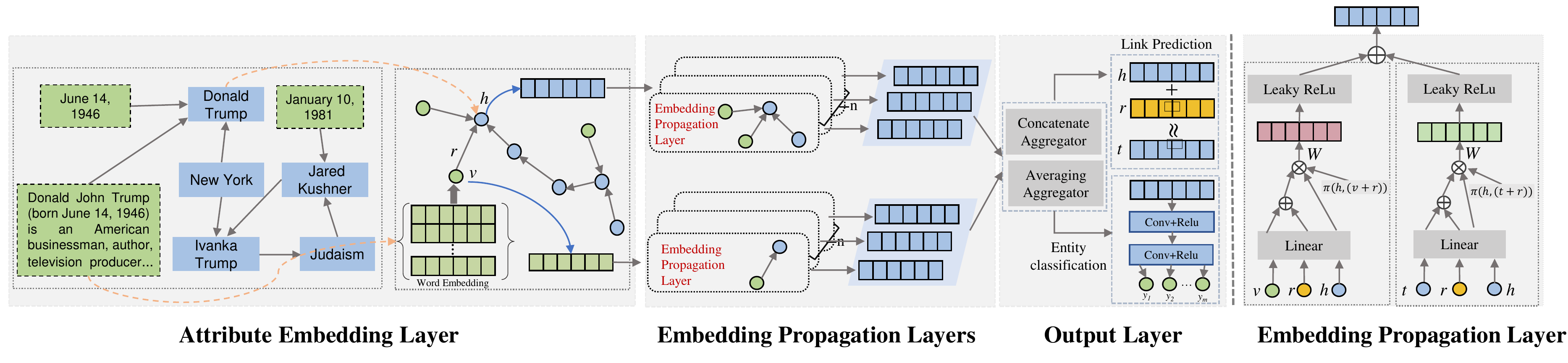}
	\caption{Illustration of the KANE architecture.}
	\label{Fig:kgs}
\end{figure*}

\subsection{Overall Architecture}

The process of KANE is illustrated in Figure \ref{Fig:kgs}. We introduce the architecture of KANE from left to right. As shown in Figure \ref{Fig:kgs}, the whole triples of knowledge graph as input. The task of attribute embedding lays is embedding every value in attribute triples into a continuous vector space while preserving the semantic information. To capture both high-order structural information of KGs, we used an attention-based embedding propagation method. This method can recursively propagate the embeddings of entities from an entity's neighbors, and aggregate the neighbors with different weights.  The final embedding of entities, relations and values are feed into two different deep neural network for two different tasks including link predication and entity classification.
\subsection{Attribute Embedding Layer}
The value in attribute triples usually is sentence or a word. To encode the representation of value from its sentence or word, we need to encode the variable-length sentences to a fixed-length vector. In this study, we adopt two different encoders to model the attribute value.

\textbf{Bag-of-Words Encoder}. The representation of attribute value can be generated by a summation of all words embeddings of values. We denote the attribute value $a$ as a word sequence $a = w_{1},...,w_{n}$, where $w_{i}$ is the word at position $i$. The embedding of $\textbf{a}$ can be defined as follows.
\begin{equation}\label{eq:grpah} 
\textbf{a}=\sum_{i=1}^{n}  \textbf{w}_{i},
\end{equation}
where $\textbf{w}_{i}\in \mathbb{R}^{k}$ is the word embedding of $w_{i}$.

Bag-of-Words Encoder is a simple and intuitive method, which can capture the relative importance of
words. But this method suffers in that two strings that contains the same words with different order will have the same representation.

\textbf{LSTM Encoder}. In order to overcome the limitation of Bag-of-Word encoder, we consider using LSTM networks to encoder a sequence of words in attribute value into a single vector. The final hidden state of the LSTM networks is selected as a representation of the attribute value.

\begin{equation}\label{eq:grpah} 
\textbf{a}=f_{lstm}(\textbf{w}_{1},\textbf{w}_{2},\textbf{w}_{3},...,\textbf{w}_{n}),
\end{equation}
where $f_{lstm}$ is the LSTM network.
\subsection{Embedding Propagation Layer}
Next we describe the details of recursively embedding propagation method building upon the architecture of graph convolution network. Moreover, by exploiting the idea of graph attention network, out method learn to assign varying levels of importance to entity in every entity's neighborhood and can generate attentive weights of cascaded embedding propagation. In this study, embedding propagation layer consists of two mainly components: attentive embedding propagation and embedding aggregation. Here, we start by describing the attentive embedding propagation.

\textbf{Attentive Embedding Propagation:} Considering an KG $G$, the input to our layer is a set of entities, relations and attribute values embedding. We use $\textbf{h}\in \mathbb{R}^{k}$ to denote the embedding of entity $h$. The neighborhood of entity $h$ can be described by $\mathcal{N}_{h} = \{t,a|(h,r,t)\in T_{R} \cup (h,r,a)\in T_{A}\}$. The purpose of attentive embedding propagation is encode $\mathcal{N}_{h}$ and output a vector $\vec{\textbf{h}}$ as the new embedding of entity $h$.

In order to obtain sufficient expressive power, one learnable linear transformation $\textbf{W}\in \mathbb{R}^{k^{'} \times k}$ is adopted to transform the input embeddings into higher level feature space. In this study, we take a triple $(h,r,t)$ as example and the output a vector $\vec{\textbf{h}}$ can be formulated as follows:

\begin{equation}\label{eq:neigh} 
\vec{\textbf{h}} = \sum_{t \in \mathcal{N}_{h} } \pi(h,r,t) \textbf{W} (\textbf{r} +\textbf{t} ),
\end{equation}

where $\pi(h,r,t)$ is attention coefficients which indicates the importance of entity's $t$ to entities $h$ .

In this study, the attention coefficients also control how many information being propagated from its neighborhood through the relation. To make attention coefficients easily comparable between different entities, the attention coefficient $\pi(h,r,t)$ can be computed using a \textit{softmax function} over all the triples connected with $h$. The softmax function can be formulated as follows:
\begin{equation}\label{eq:coefficient} 
\pi(h,r,t) = \frac{exp(\pi(h,r,t))}{ \sum_{t' \in \mathcal{N}_{h} } exp(\pi(h,r',t'))}.
\end{equation}

Hereafter, we implement the attention coefficients $\pi(h,r,t)$ through a single-layer feedforward neural network, which is formulated as follows:
\begin{equation}\label{eq:attention} 
\pi(h,r,t) = LeakyRelu((\textbf{W}\textbf{r})^{T} \textbf{W} (\textbf{r} +\textbf{t} )),
\end{equation}
where the leakyRelu is selected as activation function. 

As shown in Equation \ref{eq:attention}, the attention coefficient score is depend on the distance head entity $h$ and the tail entity $t$ plus the relation $r$, which follows the idea behind TransE that the embedding $\textbf{t}$ of head entity should be close to the tail entity's embedding $\textbf{r}$ plus the relation vector $\textbf{t}$ if $(h, r, t)$ holds. 

\textbf{Embedding Aggregation}. To stabilize the learning process of attention, we perform multi-head attention on final layer. Specifically, we use $m$ attention mechanism to execute the transformation of Equation \ref{eq:neigh}. A aggregators is needed to combine all embeddings of multi-head graph attention layer. In this study, we adapt two types of aggregators:
\begin{itemize}
	\item Concatenation Aggregator concatenates all embeddings of multi-head graph attention, followed by a nonlinear transformation:
	\begin{equation}\label{eq:Linear1} 
	\vec{\textbf{h'}} = LeakyReLu(\textbf{W}(\mathop{\Big|\Big|}\limits_{i=1}^{m} \sum_{t \in \mathcal{N}_{h} } \pi(h,r,t)^{i} \textbf{W}^{i} (\textbf{r} +\textbf{t}))),
	\end{equation}
	where $\mathop{\Big|\Big|}$ represents concatenation, $ \pi(h,r,t)^{i}$ are normalized attention coefficient computed by the $i$-th attentive embedding propagation, and $\textbf{W}^{i}$ denotes the linear transformation of input embedding.
	\item Averaging Aggregator sums all embeddings of multi-head graph attention and the output embedding in the final is calculated applying averaging:
	\begin{equation}\label{eq:Linear1} 
	\vec{\textbf{h'}} = LeakyReLu(\frac{1}{m}(\sum\limits_{i=1}^{m} \sum_{t \in \mathcal{N}_{h} } \pi(h,r,t)^{i} \textbf{W}^{i} (\textbf{r} +\textbf{t}))).
	\end{equation}
\end{itemize}
In order to encode the high-order connectivity information in KGs, we use multiple embedding propagation layers to gathering the deep information propagated from the neighbors. More formally, the embedding of entity $h$ in $l$-th layers can be defined as follows:
\begin{equation}\label{eq:Linear1} 
\vec{\textbf{h}}^{(l)} = \sum_{t \in \mathcal{N}_{h} } \pi(h,r,t) \textbf{W} (\textbf{r}^{(l-1)} +\textbf{t}^{(l-1)}).
\end{equation}
After performing $L$ embedding propagation layers, we can get the final embedding of entities, relations and attribute values, which include both high-order structural and attribute information of KGs. Next, we discuss the loss functions of KANE for two different tasks and introduce the learning and optimization detail. 

\subsection{Output Layer and Training Details}
Here, we introduce the learning and optimization details for our method. Two different loss functions are carefully designed fro two different tasks of KG, which include knowledge graph completion and entity classification. Next details of these two loss functions are discussed. 

\textbf{knowledge graph completion}. This task is a classical task in knowledge graph representation learning community. Specifically, two subtasks are included in knowledge graph completion: entity predication and link predication. Entity predication aims to infer the impossible head/tail entities in testing datasets when one of them is missing, while the link predication focus on complete a triple when relation is missing. In this study, we borrow the idea of translational scoring function from TransE, which the embedding $\textbf{t}$ of tail entity should be close to the head entity's embedding $\textbf{r}$ plus the relation vector $\textbf{t}$ if $(h, r, t)$ holds, which indicates $d(h+r,t)= ||\textbf{h}+\textbf{r}- \textbf{t}||$. Specifically, we train our model using hinge-loss function, given formally as

\begin{equation}\label{eq:Linear1} 
\mathcal{L}=\sum_{(h,r,e) \in T \ }\sum_{(h',r,e') \in T'} \lbrack \gamma +d(h+r,e)-d(h'+r-e') \rbrack_{+},
\end{equation}
where $\gamma>0$ is a margin hyper-parameter, $\lbrack x \rbrack_{+}$ denotes the positive part of $x$, $T=T_{R} \cup T_{A}$ is the set of valid triples, and $T'$ is set of corrupted triples which can be formulated as:

\begin{equation} 
T'=\{(h',r,e)| h' \in E \} \cup \{(h,r,e')| e' \in E \}.
\end{equation}

\textbf{Entity Classification}. For the task of entity classification, we simple uses a fully connected layers and binary cross-entropy loss (BCE) over sigmoid activation on the output of last layer. We minimize the binary cross-entropy on all labeled entities, given formally as:

\begin{equation}\label{eq:Linear1} 
\mathcal{L}=-\frac{1}{|E_{D}|} \sum_{e \in E_{D}} \sum_{j=1}^{C}\lbrack y_{ej}\log(\sigma(f_{ej}))+(1-y_{ej})\log(1-\sigma(f_{ej})) \rbrack
\end{equation}
where $E_{D}$ is the set of entities indicates have labels, $C$ is the dimension of the output features, which is equal to the number of classes, $y_{ej}$ is the label indicator of entity $e$ for $j$-th class, and $\sigma(x)$ is sigmoid function $\sigma(x) = \frac{1}{1+e^{-x}}$.

We optimize these two loss functions using mini-batch stochastic gradient decent (SGD) over the possible $\textbf{h}$,  $\textbf{r}$, $\textbf{t}$, with the chin rule that applying to update all parameters. At each step, we update the parameter $\textbf{h}^{\tau+1}\leftarrow\textbf{h}^{\tau}-\lambda\nabla_{\textbf{h}}\mathcal{L}$, where $\tau$ labels the iteration step and $\lambda$ is the learning rate.

\section{Experiments}
\subsection{Date sets} 
In this study, we evaluate our model on three real KG including two typical large-scale knowledge graph: Freebase \cite{bollacker2008freebase}, DBpedia \cite{auer2007dbpedia} and a self-construction game knowledge graph. First, we adapt a dataset extracted from Freebase, i.e., \textbf{FB24K}, which used by \cite{lin2016knowledge}. Then, we collect extra entities and relations that from DBpedia which that they should have at least 100 mentions \cite{bordes2013translating} and they could link to the entities in the \textbf{FB24K} by the sameAs triples. Finally, we build a datasets named as \textbf{DBP24K}. In addition, we build a game datasets from our game knowledge graph, named as \textbf{Game30K}. The statistics of datasets are listed in Table \ref{table:statcis}.
\begin{table}[!htp]
	\caption{The statistics of datasets.}\smallskip
	\centering
	\setlength{\tabcolsep}{3mm}{
		\begin{tabular}{l|l|l|l}
			\hline
			Datasets & FB24K & DBP24K & Game30K \\ \hline
			\#Entities & 23,643 &22,951  & 30,845  \tabularnewline
			\#Relations & 673 & 2,561 & 1,318   \tabularnewline
			\#Attributes & 314 & 1,561 & 2,760   \tabularnewline
			\#Total Triples & 423,560 & 437,561 &370,140   \tabularnewline
			\hline
		\end{tabular}
	}
	\label{table:statcis}
\end{table}

\begin{table*}[htbp]
	\centering
	\caption{Entity classification results in accuracy. We run all models 10 times and report mean $\pm$ standard deviation. KANE significantly outperforms baselines on FB24K, DBP24K and Game30K.}\smallskip
	\label{table:classification}
	\begin{tabular}{l|l|c|c|c}
		\hline
		Types & Methods  & FB24K  & DBP24K  & Game30K                               \\
		\hline
		\multirow{3}{*}{Typical} & TransE + LR   &   0.5819 $\pm 0.0015 $        & 0.6124 $\pm 0.0001 $        &    0.6315 $\pm 0.0018 $      \tabularnewline
		
		&TransR + LR    &   0.6012 $\pm 0.0017 $        & 0.6516 $\pm 0.0018 $        &    0.6731 $\pm 0.0026 $            \tabularnewline
		&TransH + LR   &   0.6129 $\pm 0.0005 $        & 0.6439 $\pm 0.0054 $        &    0.6821 $\pm 0.0052 $                       \tabularnewline
		\hline
		\multirow{2}{*}{Path-based } & PTransE + LR    &   0.7564 $\pm 0.0031 $        & 0.8119 $\pm 0.0031 $        &    0.8041 $\pm 0.0000 $      \tabularnewline
		
		&ALL-PATHS + LR    &    0.7625 $\pm 0.0037 $        & 0.8155 $\pm 0.0041 $        &    0.8172 $\pm 0.0049 $             \tabularnewline
		\hline
		\multirow{2}{*}{Attribute-incorporated } & KR-EAR + LR   &   0.7319 $\pm 0.0004 $        & 0.7962 $\pm 0.0005 $        &    0.8092 $\pm 0.0062 $      \tabularnewline
		
		&R-GCN + LR    &   0.7721 $\pm 0.0022 $        & 0.8193 $\pm 0.0041 $        &    0.8229 $\pm 0.0054 $             \tabularnewline
		\hline
		\multirow{4}{*}{Our Methods} & KANE (BOW+Concatenation)    &   0.7852 $\pm 0.0013 $        & 0.8205 $\pm 0.0012 $        &    0.8312 $\pm 0.0008 $      \tabularnewline
		
		& KANE (BOW+Average)    &   0.7745 $\pm 0.0015 $        & 0.8221 $\pm 0.0095 $        &    0.8293 $\pm 0.0037 $            \tabularnewline
		& KANE (LSTM+Concatenation)    &   \textbf{0.8011} $\pm \textbf{0.0011} $        & \textbf{0.8592} $\pm \textbf{0.0062} $        &    \textbf{0.8605} $\pm  \textbf{0.0033} $                       \tabularnewline
		& KANE (LSTM+Average)    &   0.7929 $\pm 0.0018 $        & 0.8236 $\pm 0.0021 $        &    0.8523 $\pm 0.0031 $                      \tabularnewline
		\hline
	\end{tabular}
	
\end{table*}

\subsection{Experiments Setting} In evaluation, we compare our method with three types of models:

1) Typical Methods. Three typical knowledge graph embedding methods includes TransE, TransR and TransH are selected as baselines. For TransE, the dissimilarity measure is implemented with L1-norm, and relation as well as entity are replaced during negative sampling. For TransR, we directly use the source codes released in \cite{lin2015learning}. In order for better performance, the replacement of relation in negative sampling is utilized according to the suggestion of author. 

2) Path-based Methods. We compare our method with two typical path-based model include PTransE, and ALL-PATHS \cite{toutanova2016compositional}. PTransE is the first method to model relation path in KG embedding task, and ALL-PATHS improve the PTransE through a dynamic programming algorithm which can incorporate all relation paths of bounded length.

3) Attribute-incorporated Methods. Several state-of-art attribute-incorporated methods including R-GCN \cite{schlichtkrull2018modeling} and KR-EAR \cite{lin2016knowledge} are used to compare with our methods on three real datasets.

In addition, four variants of KANE which each of which correspondingly defines its specific way of computing the attribute value embedding and embedding aggregation are used as baseline in evaluation. In this study, we name four three variants as KANE (BOW+Concatenation), KANE (BOW+Average), and  KANE (LSTM+Concatenation), KANE (LSTM+Average). Our method is learned with mini-batch SGD. As for hyper-parameters, we select batch size among \{16, 32, 64, 128\}, learning rate $\lambda$ for SGD among \{0.1, 0.01, 0.001\}. For a fair comparison, we also set the vector dimensions of all entity and relation to the same $k \in $\{128, 258, 512, 1024\}, the same dissimilarity measure $l_{1}$ or $l_{2}$ distance in loss function, and the same number of negative examples $n$ among \{1, 10, 20, 40\}. The training time on both data sets is limited to at most 400 epochs. The best models are selected by a grid search and early stopping on validation sets.

\subsection{Entity Classification}
\subsubsection{Evaluation Protocol.} In entity classification, the aim is to predicate the type of entity. For all baseline models, we first get the entity embedding in different datasets through default parameter settings as
in their original papers or implementations.Then, Logistic Regression is used as classifier, which regards the entity's embeddings as feature of classifier. In evaluation, we random selected 10\% of training set as validation set and accuracy as evaluation metric.

\subsubsection{Test Performance.} Experimental results of entity classification on the test sets of all the datasets is shown in Table \ref{table:classification}. The results is clearly demonstrate that our proposed method significantly outperforms state-of-art results on accuracy for three datasets. For more in-depth performance analysis, we note: (1)  Among all baselines, Path-based methods and Attribute-incorporated methods outperform three typical methods. This indicates that incorporating extra information can improve the knowledge graph embedding performance; (2) Four variants of KANE always outperform baseline methods. The main reasons why KANE works well are two fold: 1) KANE can capture high-order structural information of KGs in an efficient, explicit manner and passe these information to their neighboring; 2) KANE leverages rich information encoded in attribute triples. These rich semantic information can further improve the performance of knowledge graph; (3) The variant of KANE that use LSTM Encoder and Concatenation aggregator outperform other variants. The main reasons is that LSTM encoder can distinguish the word order and concatenation aggregator combine all embedding of multi-head attention in a higher leaver feature space, which can obtain sufficient expressive power.
\begin{figure}[!t]
	\centering
	\subfloat[DBP24K]{
		\includegraphics[width=0.25\textwidth]{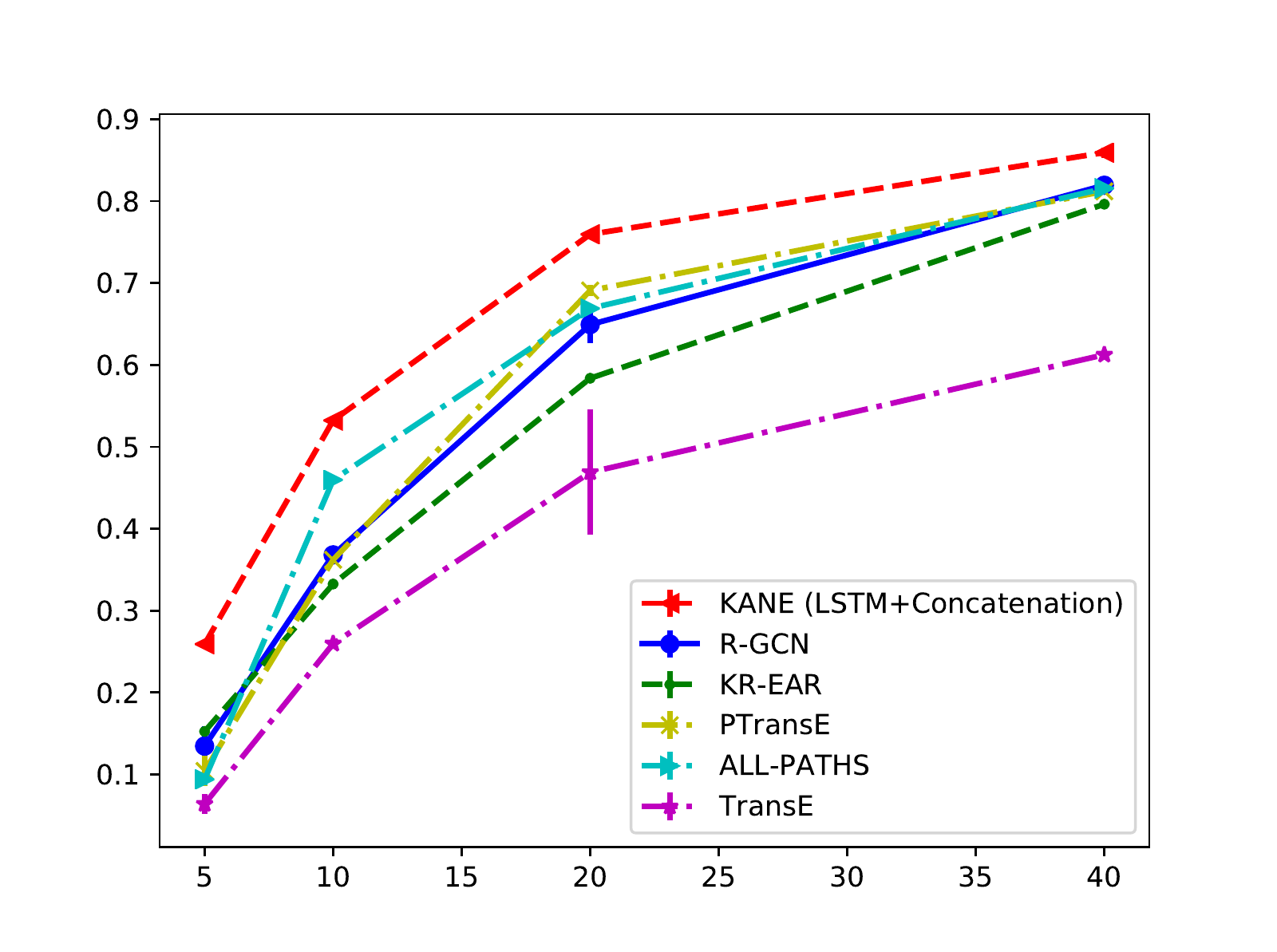}}
	\subfloat[Game24K]{
		\includegraphics[width=0.25\textwidth]{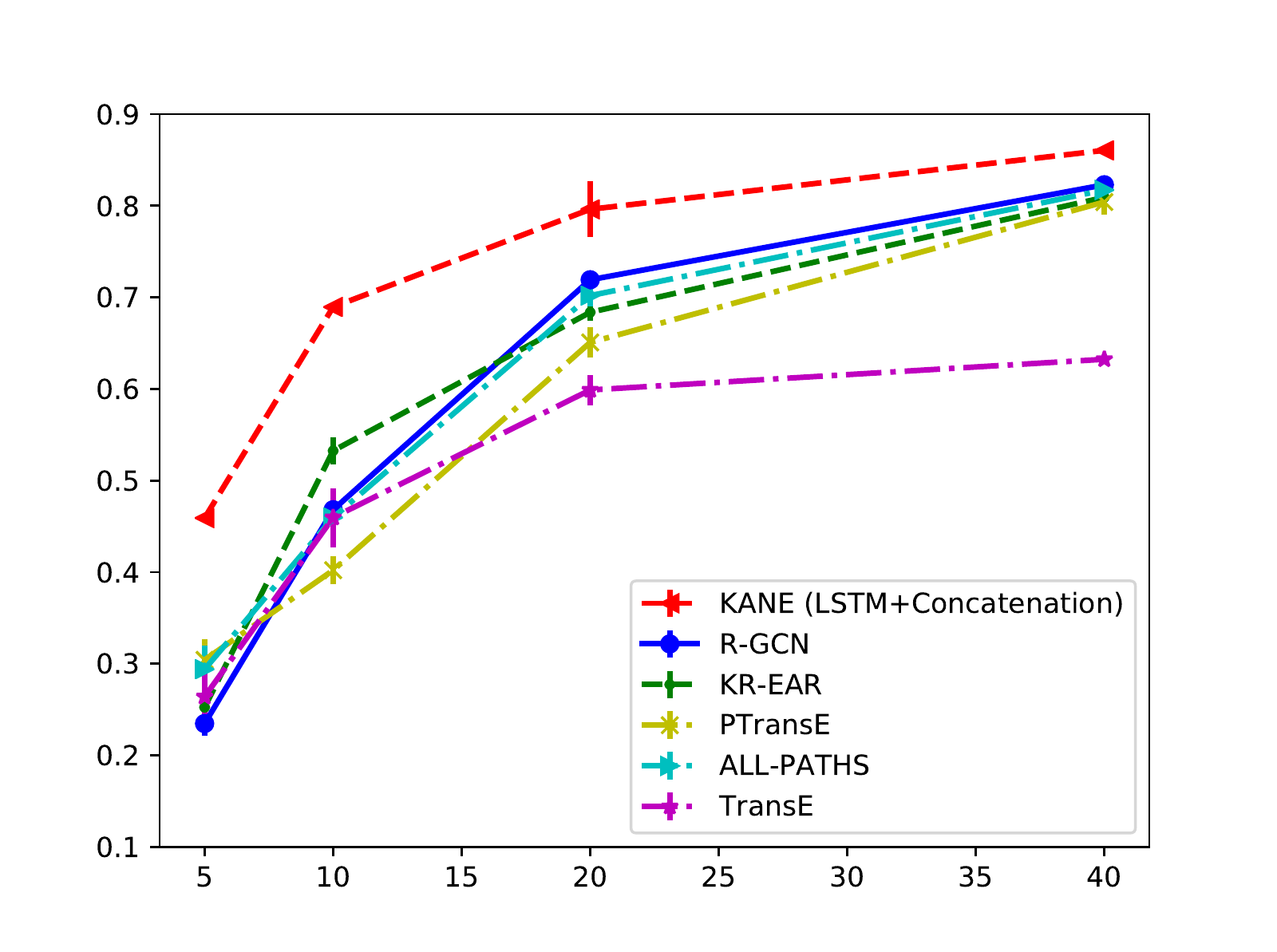}}
	\caption{Test accuracy with increasing epoch.}
	\label{Fig:epoch}
\end{figure} 
\begin{figure}[!t]
	\centering
	\subfloat[Embedding size.]{
		\includegraphics[width=0.258\textwidth]{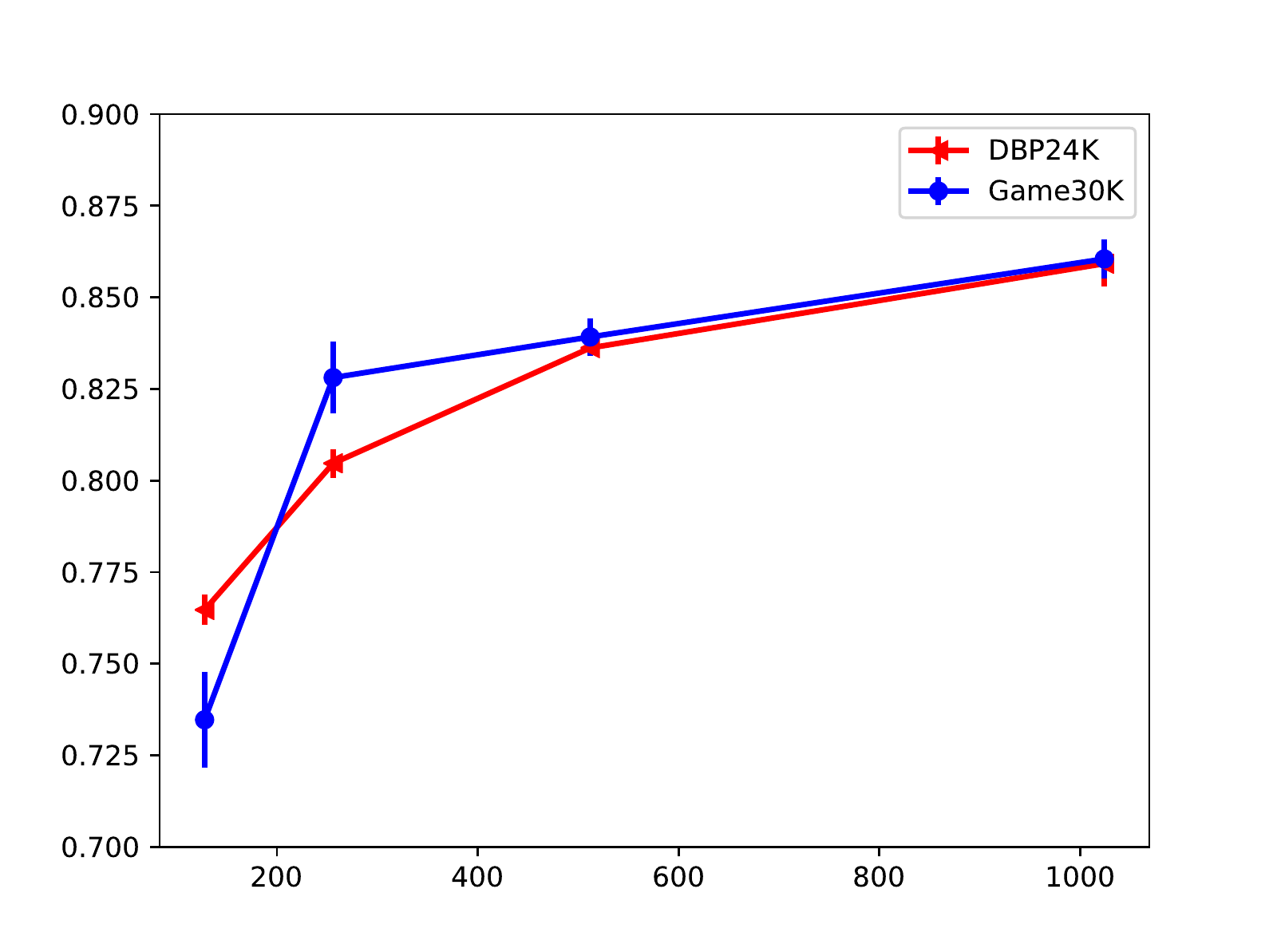}}
	\subfloat[Training data proportions]{
		\includegraphics[width=0.25\textwidth]{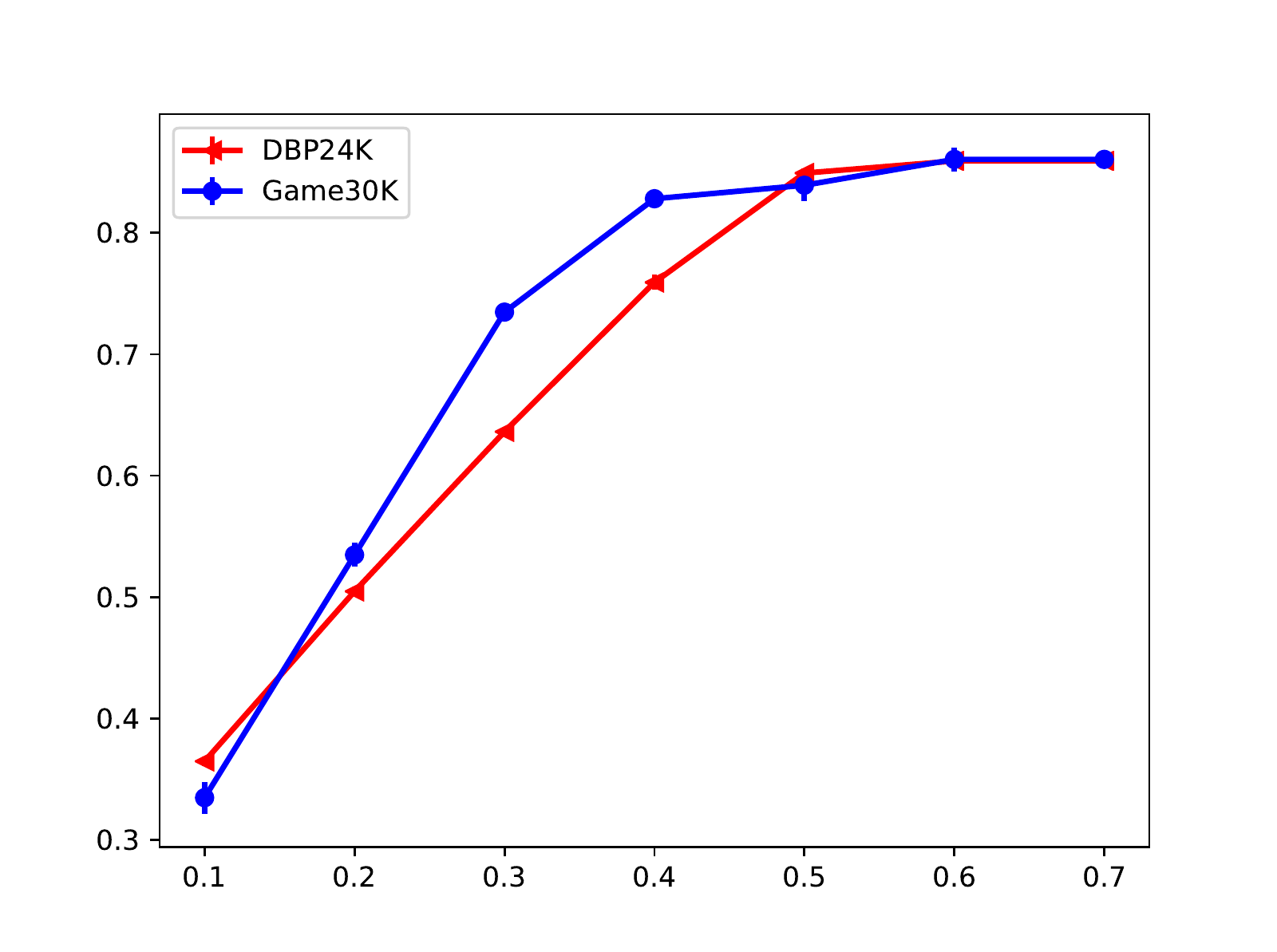}}
	\caption{Test accuracy by varying parameter.}
	\label{Fig:parameter}
\end{figure} 
\begin{figure}[!t]
	\centering
	\subfloat[R-GCN]{
		\includegraphics[width=0.24\textwidth]{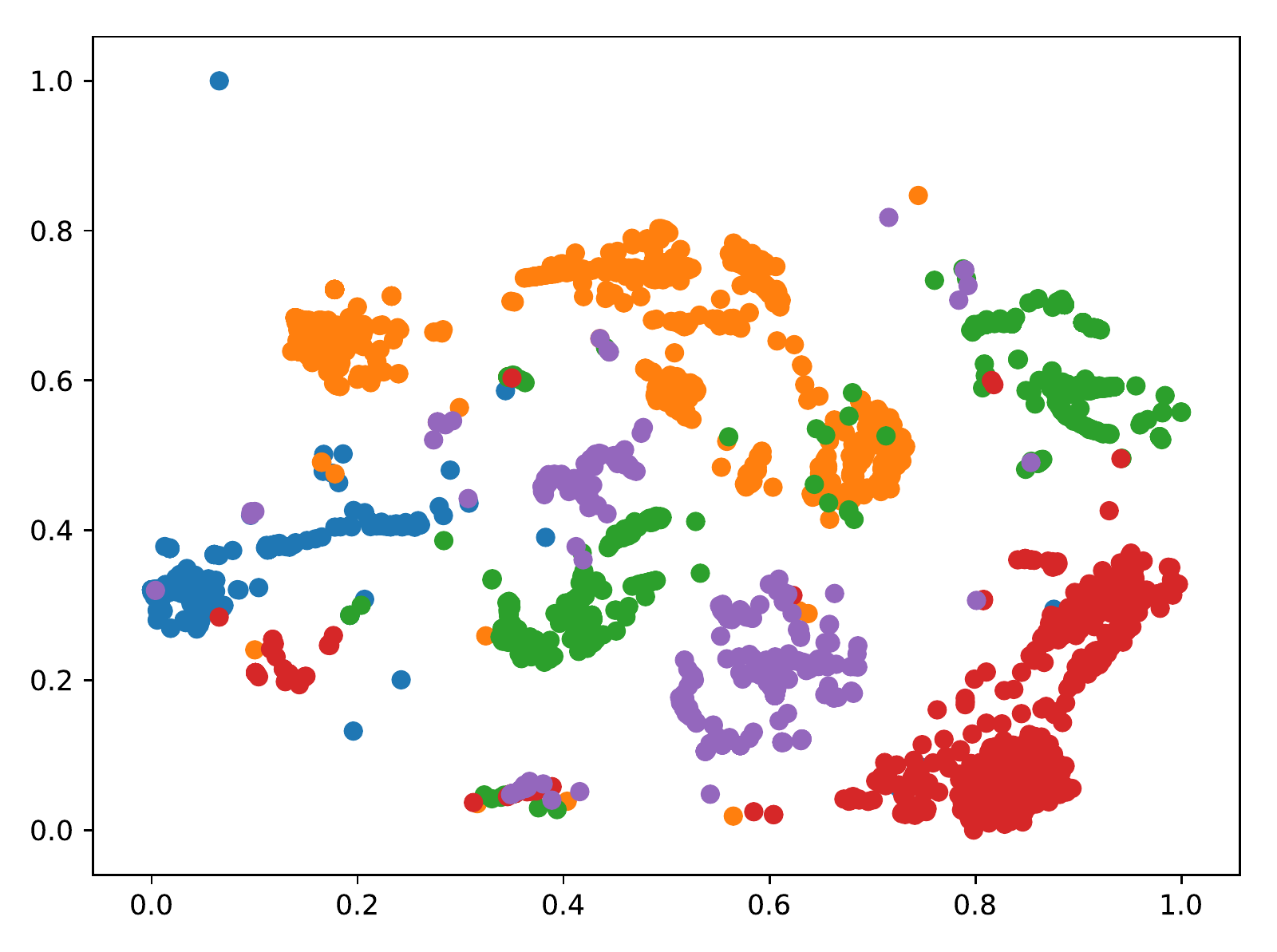}}
	\subfloat[KANE]{
		\includegraphics[width=0.24\textwidth]{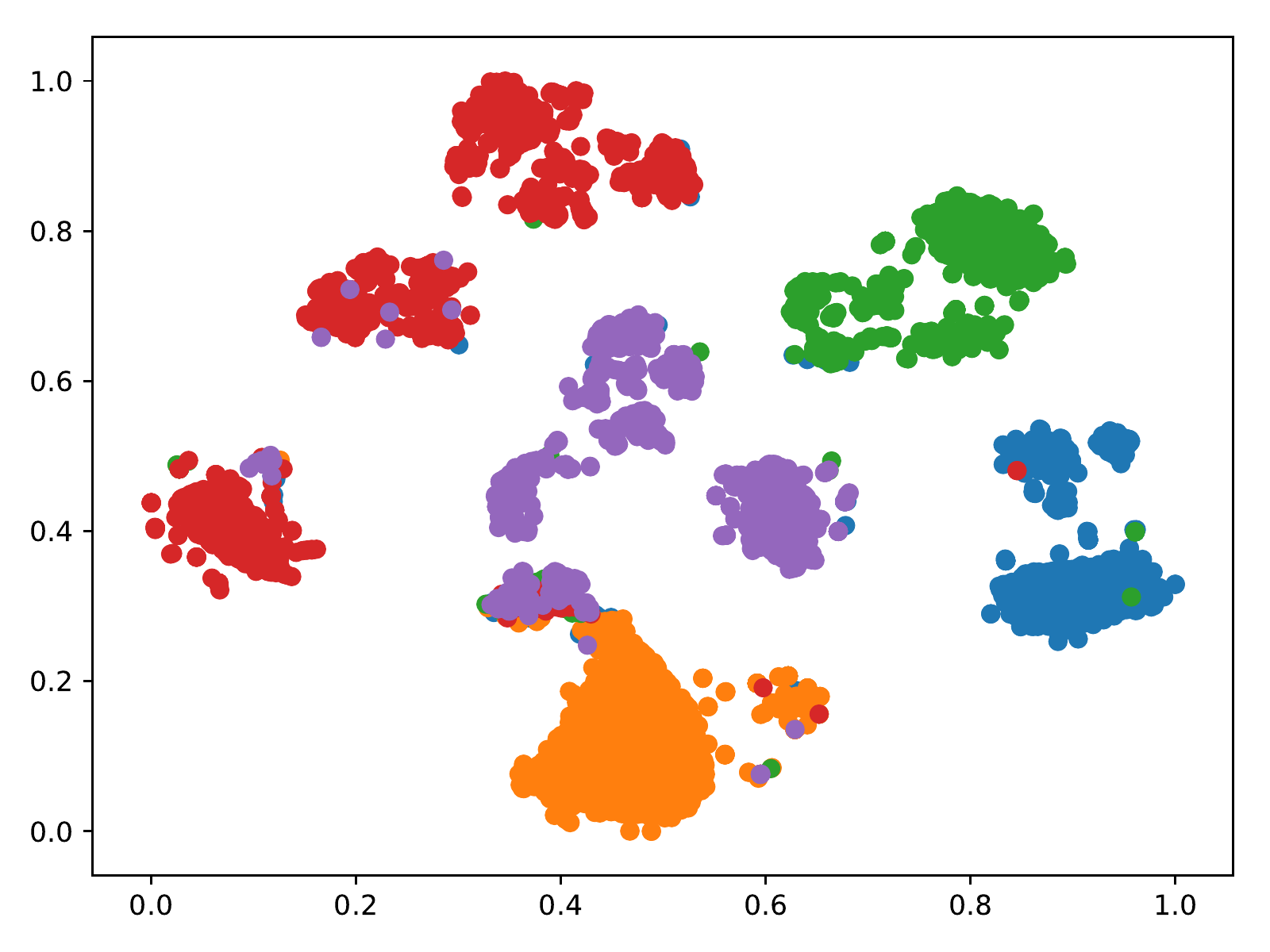}}\\
	\subfloat[TransE]{
		\includegraphics[width=0.24\textwidth]{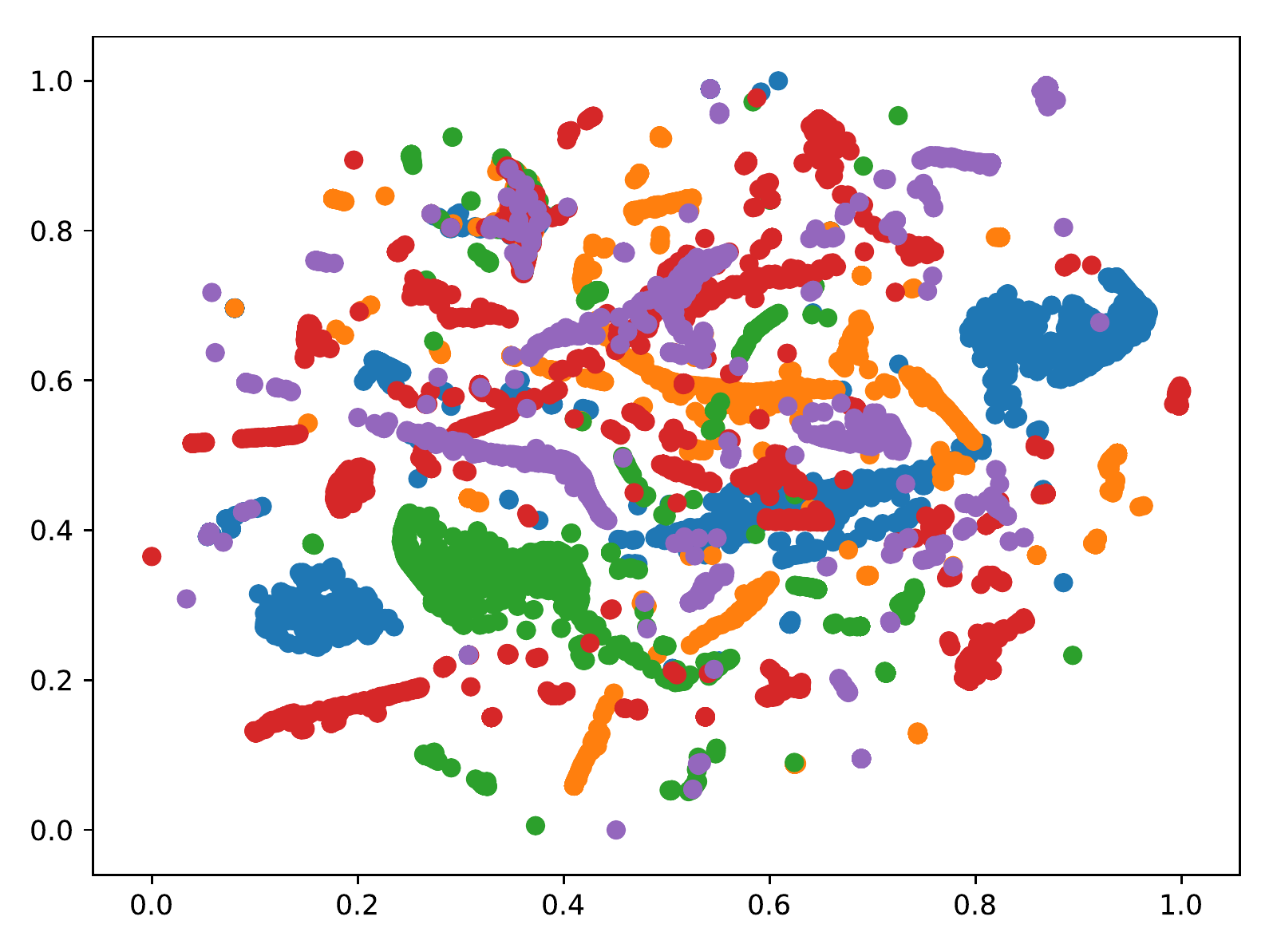}}
	\subfloat[PransE]{
		\includegraphics[width=0.24\textwidth]{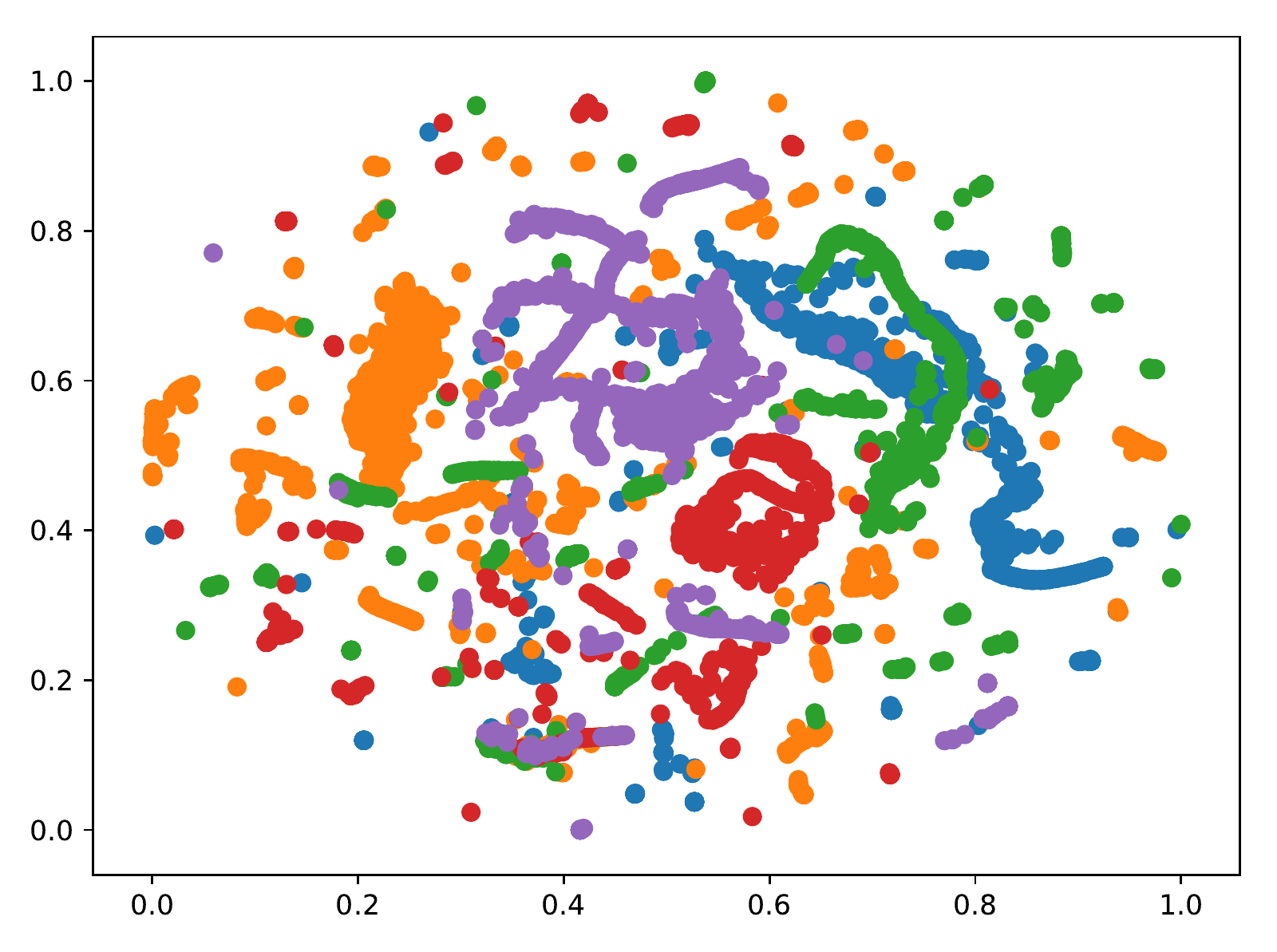}}
	\caption{The t-SNE visualization of entity embeddings in Game30K.}
	\label{Fig:visualization}
\end{figure} 
\begin{table*}[!t]
	\centering
	\caption{Results of knowledge graph completion (FB24K)}
	\label{table:completion}
	\begin{tabular}{c|cc|cc|cc|cc}
		\hline
		\multirow{3}{*}{Metric} & \multicolumn{4}{c|}{Entity Predication} & \multicolumn{4}{c}{Relation Predication} \tabularnewline
		& \multicolumn{2}{c|}{Hits@10} &  \multicolumn{2}{c|}{Mean Rank} &  \multicolumn{2}{c|}{Hits@1} &  \multicolumn{2}{c}{Mean Rank} \tabularnewline
		& Raw&Filter& Raw&Filter& Raw&Filter& Raw&Filter\tabularnewline
		\hline
		TransE &35.8  &53.0  &259  & 200 &65.9&83.8 &3.1 &  2.8 \tabularnewline
		TransR& 37.0& 56.1 &260&200   & 65.2&84.5  & 3.4& 3.1 \tabularnewline
		TransH & 33.9 & 50.2 &282  & 224 &64.9 &84.1 &3.4 &  3.1  \tabularnewline
		\hline
		PTransE  & 37.3 & 56.1 &249  & 131 &67.3  &86.1 &2.4 &2.1  \tabularnewline 
		ALL-PATHS  & 38.6 &59.9  & 208 & 121 & 67.8 & 86.4 &2.1 &1.9  \tabularnewline
		\hline
		KR-EAR  & 38.5 & 54.5 &186  & 133 &67.9  &86.2 &2.4 &2.1  \tabularnewline 
		R-GCN  & \textbf{42.3} &59.1  & \textbf{151} & 119 & 68.8 & 87.2 &2.1 &2.0  \tabularnewline
		\hline
		KANE (BOW+Concatenation) & 37.4 &57.4  & 201 & 123& 68.9 & 80.1&2.2 &2.1  \tabularnewline
		KANE (BOW+Average) & 34.3 &55.8  & 189 & 131 &68.1 &87.3&2.4 & 2.2\tabularnewline
		KANE (LSTM+Concatenation) & 41.5 & \textbf{61.2}  & 162 & \textbf{103} & \textbf{69.4} & \textbf{88.1}& \textbf{1.9} & \textbf{1.8} \tabularnewline
		KANE (LSTM+Average) & 34.3 &59.8  & 173 & 108 &69.1  &87.3&2.2 & 2.1\tabularnewline
		\hline
	\end{tabular}
\end{table*}
\subsubsection{Efficiency Evaluation.} Figure \ref{Fig:epoch} shows the test accuracy with increasing epoch on DBP24K and Game30K. We can see that test accuracy first rapidly increased in the first ten iterations, but reaches a stable stages when epoch is larger than 40. Figure \ref{Fig:parameter} shows test accuracy with different embedding size and training data proportions. We can note that too small embedding size or training data proportions can not generate sufficient global information. In order to further analysis the embeddings learned by our method, we use t-SNE tool \cite{maaten2008visualizing} to visualize the learned embedding. Figure \ref{Fig:visualization} shows the visualization of 256 dimensional entity's embedding on Game30K learned by KANE, R-GCN, PransE and TransE. We observe that our method can learn more discriminative entity's embedding than other other methods.

\subsection{Knowledge Graph Completion}
The purpose of knowledge graph completion is to complete a triple $(h, r, t)$ when one of $h, r, t$ is missing, which is used many literature \cite{bordes2013translating}. Two measures are considered as our evaluation metrics: (1) the mean rank of correct entities or relations (Mean Rank); (2) the proportion of correct entities or relations ranked in top1 (Hits@1, for relations) or top 10 (Hits@10, for entities). Following the setting in \cite{bordes2013translating}, we also adopt the two evaluation settings named "raw" and "filter" in order to avoid misleading behavior.

The results of entity and relation predication on FB24K are shown in the Table \ref{table:completion}. This results indicates that KANE still outperforms other baselines significantly and consistently. This also verifies the necessity of modeling high-order structural and attribute information of KGs in Knowledge graph embedding models.
\section{Conclusion and Future Work}
Many recent works have demonstrated the benefits of knowledge graph embedding in knowledge graph completion, such as relation extraction. However, We argue that knowledge graph embedding method still have room for improvement. First, TransE and its most extensions only take direct relations between entities into consideration. Second, most existing knowledge graph embedding methods just leverage relation triples of KGs while ignoring a large number of attribute triples. In order to overcome these limitation, inspired by the recent developments of graph convolutional networks, we propose a new knowledge graph embedding methods, named KANE. The key ideal of KANE is to aggregate all attribute triples with bias and perform embedding propagation based on relation triples when calculating the representations of given entity. Empirical results on three datasets show that KANE significantly outperforms seven state-of-arts methods.

\bibliographystyle{aaai}\bibliography{mgtransE}
\end{document}